# A machine-learning software-systems approach to capture social, regulatory, governance, and climate problems

Dr. Christopher A. Tucker

Cartheur Robotics, Prague, Czechia

cartheur@protonmail.com

**Abstract**

This paper will discuss the role of an artificially-intelligent computer system as critique-based, implicit-organizational, and an inherently necessary device, deployed in synchrony with parallel governmental policy, as a genuine means of capturing nation-population complexity in quantitative form, public contentment in societal-cooperative economic groups, regulatory proposition, and governance-effectiveness domains. It will discuss a solution involving a well-known algorithm and proffer an improved mechanism for knowledge-representation, thereby increasing range of utility, scope of influence (in terms of differentiating class sectors) and operational efficiency. It will finish with a discussion of these and other historical implications.

**Introduction**

The world created by humans to manage their daily affairs is growing in complexity beyond the comprehension capability of the vast majority of them. The political classes are vulnerable to implementation of policy that proves incorrect and damages the credibility of the state over the long term. Some persons are keenly aware of this matter and are proposing solutions scrutinized by arbitrarily-organized armies of developers, data scientists, software architects, and application engineers. While sounding good in theory, it is a dangerous activity to pursue over time as complexities and coordination between the parts of the teams become unmanageable. A better strategy is to shadow these efforts with a learning algorithm manifest in its own hardware acting as implementation-in-chief to assess the effectiveness of the given policy application, as well as the organizational efforts of the teams. This paper will discuss the role of a very particular computer system as an effective organizational criterion to manage human synchrony-driven creative solutions to complex societal and governance problems assessed as imperative by the leadership of the state.

At first it might seem surprising that society and governance lies not only with national governments, diplomatic attaches, and international agreements but also with the goals and ambitions of commerce. This is because of the adversarial nature of civilization as it has evolved from its beginnings in ancient times to the present day. Persons with vested interests in one area of the world necessarily need persons from another area of the world not only for the reasons of trade and commerce, but in terms intellectual capacity as well. This is not liberalism but rationalism in a global context. Stakeholders in companies require expertise from stakeholders other companies, requiring labor and skills that take too much time and expense for them to develop independently.

**The problem**

So lies the conundrum today manifest as an irony between despite an increasing connectivity and communication globally, trappings by unwise policy and unwavering faith in status quo leads to disdain, isolationism, and vilification of persons offering alterative interpretations of problems facing societal activities on increasingly larger scales. Over long and short periods of time, it can lead to corrosion in relations between countries as the idea of an imperialistic imperative in functioning capitalistic-democratic populations extrude their unconscious wants and desires yielding the singular purpose of civilization: to find negotiated solutions in lieu of violent conflict. Since antiquity, parties from disparate cultures have used their words to find common ground, sending armies of diplomats and advisors scribbling furiously to capture the essence of desires of the parties in order the solution benefit both in maximum capacity. This exercise has proven futile. Failure of governments worldwide to capture correctly the needs of their citizens, including frames of context, is the key reason for global unrest. The failure is plainly visible by those exercising power not having the tools to understand how their governmental system is functioning *in real time* by seeing the state of affairs, including algedonic feedback of its citizen's activity, without resorting to Orwellian memes.

Can the political classes be blamed? Certainly, this is the first response. However, does this provide the proper motivation to correct the steerage of the day-to-day business of society and commerce in order that it satisfies the public's needs? Probably not, but what other mechanisms are there? In the end, it is people running the affairs of state and it is people inhabiting its borders and traversing its boundaries. No improvement can neither be gleaned nor understood as the correct method and means to finding tangible solutions.

Are there technological solutions available? The answer is both 'yes' and 'no' without prejudice toward a grey area lying between. While we currently enjoy advances in computer and quality of manufacturing, automation to reduce carbon footprints and labor costs while maintaining social contentment and thereby long-term stability of the political classes, has been a slow uptake. There are that who think that pushing further into automation with its gem, artificial intelligence, is the correct way forward. This is probably correct for the reason that our world has become complex enough that our brains can no longer reconcile it. This is simultaneously probably incorrect as the premise by which stakeholders implement A.I. is fatally flawed, as no one understands the phenomenon enough to make a proper model of it in machine form.

**Advice-by-A.I. as solution**

The solution lies in having A.I. tell us the right way forward. As petrifying as that sounds it is the only solution available to us. It is already being undertaken by governments employing armies of developers and data scientists in the hope of finding the trail of breadcrumbs leading to the solution. However, current efforts are no more than applying unsupervised (machine) learning to copious flows of data, trainspotting those parts deemed important to remember. Therefore, how such a machine is developed if:

- No one mind can comprehend the complexity of the problem, thereby,
- it cannot be properly captured, because,
- no consensus can be arrived, as,
- no one mind can comprehend *why* the model is necessary, because,
    - they not only cannot what it is designed to do, and,
    - cannot find reassurance since they are unaware of its procedural execution,
- therefore, the idea will never be explored.

The field of transparent A.I. works to reveal these dark areas of the technology. When we are enlightened to what the machine is doing, we think we are better for it and have come away with some additional meaning to apply to our lives. This is not true. It is a lie. The human mind cannot compute probability and loses resolution to comprehend complexity in groups above five members. Of the plethora of contextual information—transformed data—newly available under such would be of little value to a human. It is more important to find ways *to build trust* in an intelligent computer system. Since the public obtains its knowledge of A.I. via secondary sources, mostly in terms of entertainment, it is necessary to explore only historical examples of implemented systems, hardware, and code.

In the 1970s, a mathematician named Anthony Stafford Beer devised a project and proposed a system called *CyberSyn* in Chile under the mandate of the socialist government of Salvador Allende. CyberSyn was a project designed to implement Beer and Ashby's idea of *requisite variety* model to placing data via processes requiring further inputs by statistically distributing them in such a way as to allow counterfactual reasoning by a program executing a series of statistical equations over the time devoted to differential analysis operating against a given problem domain, in order to glean insights into how to predict its future behavior. The theory was derived from the synthesis of metaphors drawn from biology and engineering and characteristic of work in his field of cybernetics. The law of requisite variety describes that variety in the control system must match the variety in the system to be controlled. Therefore, it is interesting to attempt to apply Beer's viable systems model to control activities of actors rewriting the policies of the state to suit available opportunities to do so.

In terms of its design, CyberSyn was a naïve Bayes filter and control system that used Forester's Dynamo compiler, equivalent to today's Matlab, which ran on an IBM 360. It monitored changes in data of system variables according to a control hierarchy, called the viable systems model. Discussion of this model is out of scope of this paper but to note its effectiveness utilized in policy deployment, by Spyridopoulos, et. al.

Mathematically speaking, the algorithm that embodies the model has been demonstrated as sound by several authors, such as Ashby and Beer, along with the notion of homeostatic equilibrium of economic activity between classes. The behavior of large-scale human systems is rather manageable, given the right set of intuitions, when the computer program is designed. However, the language of its operation is often nestled in abstractions such as cooperative collaboration between different systems, noting that its key feature was that it is, as Beer describes it: "…a system that survives. It coheres; it is integral…but it has nonetheless mechanisms and opportunities to grow and learn, to evolve and to adapt."

Further technical details are beyond the scope of this paper, but are provided in the references for further reading. In terms of a general criteria-based overview, naïve Bayes has the following characteristics as advantages and disadvantages. The advantages being: It can operate on quantitative and discrete data, it is robust to isolated noise points, can ignore irrelevant values for probability estimates when decision-making on a quadratic decision boundary. However, those weaknesses to identify zero-valued conditional probabilities and dependence between features, suggests a requisite managerial architecture to compensate, such as the viable system model.

It is made to be useful to solve practical problems is by integrating a counterfactual reasoning approach into the architecture of the program, since as it is well-known, that data—its speed of access and quality of availability—determines whether the algorithm is performant enough to be advisory of answers to queries by users, relative to its environmental representation. The counterfactual approach enriches representation by introducing interrogative questions of a higher dimension. Ordinarily, as Turing first noted, we query from a representative system to compute variables, in essence, recalling them. A question like "what happened?" is one of these. A known-map is created *a posteriori* with context between its features that could quickly be out of date, depending on the type of device the algorithm is executing. Querying redraws the map from scratch, losing the contextual links from the previous map, where they are known implicitly. Adding a higher dimension to the query by introducing questions that involve cognitive introspection is an excellent technique. As an example, "what if I act?" and "what I should have done?" are two of these. The next section of the paper will attempt to bridge counterfactual planning, in terms of probabilistic reasoning, to introduce an improved version of a CyberSyn-like structure deployed in modern cloud-based systems.

**An enhancement to the mathematical model**

The software system designed by Beer, called *Cyberstride*, was a naïve Bayes classifier executing statistical equations against Dynamo, for a sufficient variety of data-points output from production facilities across the economy. This paper will discuss the mathematical aspects of a Bayes classifier with a contribution by the author of intersecting counterfactual reasoning, creating a much more sophisticated operating program. Probability, for the purposes of this paper, is defined as a quantitative measure of uncertainty state of information or event. Since we are interested in the probability of a single or set of events—what happened, the probability is conditional of what happened in an event prior. The probability of an event may depend on the occurrence or non-occurrence of another event. The dependency, written in terms of conditional probability is $P(A|B)$, which states,

- the probability that A will happen given that B already has,
- the probability to select A among B.

As:

$$P(A|B) = \frac{P(A \cap B)}{P(B)}$$
$$P(B|A)) = \frac{P(A \cap B)}{P(A)} \quad (1)$$
$$P(A \cap B) = P(B|A)P(A) = P(A|B)P(B).$$

Given the event, A, is independent from event, B, if the conditional probability is the same as the marginal probability when $P(B|A) = P(B), P(A|B) = P(A)$. We stipulate that pre-conditioned knowledge available to the system can take either *a priori and a posteriori* forms. There are three distinct kinds of probabilities, one a statement of prior probability, unconditional probabilities, and posterior probability. The first assumes knowledge of an event *a priori*, the second, a hypothesis of the state of knowledge before new data is observed. The third a kind of conditional probability about the state of knowledge after revising the outcome based on new data applied *a posteriori*. Likelihood also exists in the formula indicating a conditional probability that our observational data holds for the given hypothesis, as illustrated in Fig.1.

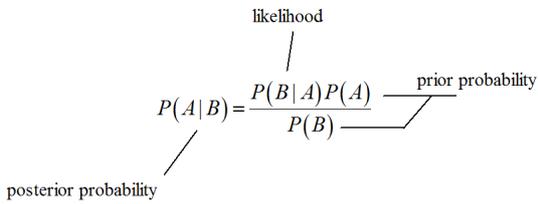

Fig.1. Bayesian predictive components.

Armed with this template, we now want to introduce hierarchical reasoning based on an apt causality. The hierarchy is broken up into three distinct levels, known as the *ladder of causation*, shown in Table 1. These components are now recombined into an effective form of the following representation. Strictly within the case of filtering-problem domains and those contextual modeling strategies, generically,

$$p(a_x|b) = \frac{p(b|a_x)p(a_x)}{p(b)}. \quad (2)$$

It is desirable to consider what the outcome would be *upon any* given assessed outcome probability. When each attribute and class label is a random variable and given a record with attributes $a_1, a_2, ..., a_n$, where the goal is to predict class $C$. Particularly, it is interesting to gain knowledge of the value of $C$ that maximizes $P(C|\sum a_n)$. An insightful approach is to computer the posterior probability $P(C|a_1, a_2, ..., a_n)$ for all values of $C$,

$$P(C \mid a_1, a_2, \ldots, a_n) = \frac{P(a_1, a_2, \ldots, a_n \mid C) P(C)}{P(a_1, a_2, \ldots, a_n)}. \tag{3}$$

When the maximum value for the conditional probability is known, choice of this value and assuming independent distributions—there is no knowledge being exchanged *between* the probabilities—it is therefore possible to apply a hierarchal causality over the transformed attributes, $P \to p$, $a_n \to x_x$, $C \to c_i$:

*What is?*
$$p(c \mid x_0) = \frac{p(x_1 x_2 \ldots x_n \mid c) p(c)}{p(x_1 x_2 \ldots x_n)}$$

*What if?*
$$p(c \mid do(x_0), y) = \frac{p(x_1 x_2 \ldots x_n \mid c) p(c)}{p(x_1 x_2 \ldots x_n)} + \frac{p(c \cap y)}{p(x_n)} \tag{4}$$

*Why?*
$$p(c \mid do(y_0), x) = \frac{p(y_1 y_2 \ldots y_n \mid c) p(c)}{p(y_1 y_2 \ldots y_n)} + \frac{p(c \cap x)}{p(y_n)}$$

$$p(c_i \mid x', y') = \prod_{k=1}^{n} \frac{p(x_1 x_2 \ldots x_n \mid c_i) p(c_i)}{p(x_1 x_2 \ldots x_n)} + \frac{p(c_i \cap y')}{p(y_n)} - \frac{p(c_i \cap x')}{p(x_n)}.$$

Keeping aware that Naïve Bayesian prediction requires each conditional probability be nonzero. Therefore, a simple additive parameter is attached, preferred to the Laplacian representation,

$$P(A_i \mid C) = \frac{N_{ic} + 1}{N_c + c}, \tag{5}$$

where $c$ is the number of classes. Adding counterfactual reasoning to a naïve Bayesian filter will substantively increase its scope and range, thereby making it far more useful on the scope of the assigned problem space. It additionally increases its environmental scope by creating known probabilistic arrangement of the data, $\frac{p(c \cap x)}{p(x)}, \frac{p(c \cap y)}{p(y)}$, expanding it to $\frac{p(c \cap x')}{p(x_m)}, \frac{p(c \cap y')}{p(y_n)}$, where the limit on knowledge of the environment is now based on the magnitude of $x_m$ and $y_n$ simultaneously.

**Discussion**

According to historical sources, Beer's design operated very well and Cyberstride proved itself successful to manage the affairs of state in late 1972, despite previous errors of the leadership to curb inflation. It demonstrated the capability of A.I., properly written and managed, could regulate and provide optimal suggestions regarding the activity of a society, given that the domain of data is available to the algorithm in that it can evolve to respond rapidly to the needs of the people. Despite its success and failures, CyberSyn was never attempted again. Persons economically threatened by A.I., those apocalyptically afraid of A.I., and those unwilling to entertain an idea that is at its very core, socialist and technocratic, oft frame it as an unpalatable solution. Certainly, the technology was developed under the mandate of a socialist government using nationalistic policy, but the government does not have to be socialist to utilize the technology at its heart. Regulated capitalistic activities and spending by the state in responding to monetary crises is also possible. It can be a conservative and centrist government as the principles of using A.I. in governance strives for peace, prosperity, and reduced expenditure because decisions are being made not only more rapidly, but also more precisely. However, the key aspect of its success is the availability of the correct data, arriving as feedback from activities of production, in order policy can be monitored as close to real time as possible, and given hardware constraints.

The solution is to construct a machine that benefits people so that they can navigate an increasingly technical world, both in terms of society and economic activity. This machine will need to be constructed even though some or most persons would not want it to be. With the advance of chaotic weather and the very real possibility we might go extinct as a species, we require such a machine to survive the coming apocalypse perhaps even by navigating it. For how can our feeble minds grasp the complexity of the weather of our planet when most of us cannot decide what to do when we are hungry, tired, angry, or amorous. Even if such prostrations of doom never materialize, the implications and their emotional weight propagate.

Climate is an important phenomenon to be keenly aware of as it has the potential of rapidly dispersing a population, introducing a high-magnitude migration event. Even if a true scientific result cannot ever be determined, monitoring of climactic events and their weight of impression upon the public will capture behaviour. This is the point: the system will communicate to the people that it understands human problems by researching solutions to present to the government whilst simultaneously not interfering in day-to-day decision-making performed by politicians. The system must be sold as functioning solely *in an advisory* capacity.

Major problems will arise in the tension between the selfish and the altruistic. The notions of democracy, what it offers, the obsession of material gain to the point malign feedback must be clarified. The rising of collaboration between left and green political classes are the most compatible way forward. As an extreme opposition, for the purposes of illustration, to counterbalance activities undertaken by other nations practicing a resurgence of Keynesian economics, however careful not to discriminate on the political spectrum between liberals, centrists, and conservatives. To avoid direct disruption to markets, counterfactual reasoning aids in more accurate leveraging of data from simulated scenarios run by the algorithm and its program beforehand. The author claims the following:

- CyberSyn is the right kind of system to operate a new kind of country, one that is as fair and equitable as possible.

- Designed in the right way and laboring under ideal philosophical constructs, such as Kant's pure reason, adding the capability to perform counterfactual reasoning, introduced by Pearl, it can accurately provide the kinds of solid predictions and proffer the right kinds of advice to the ruling political classes.

- An enhanced CyberSyn, a system specifically-designed to operate on classes of data, providing recommendations toward optimization in terms of operation, in terms of all aspects of government, policy, regulatory, trade negotiation, and even diplomacy is the *only* way forward, if it is to survive adversities while its people thrive over decades, not just over the term of the mandate.

- Current manufacturing sophistication and network connectivity is accessible to nearly all locations on the planet by utilization of non-grid-powered edge devices.

- Long term problems can be quantitatively addressed and dealt with in real time.

As time progresses, it will become painfully obvious to policy makers the dire need for A.I. and machine learning as we deal with energy, climate, commerce, infrastructure, political, human need-based services, and eventually diplomacy. The question is not of its adoption but in the manner by which it is architected and deployed. As noted earlier in this paper, if we start with the supposition that we cannot know the correct answer, trying to synthesize it with principles of equivalence instills the system to be unbiased and fair. Manipulation is a tempting supposition, but will be the rue of the design. Therefore, it is valid to insist upon an equal weight of emphasis on data independent of human introspection. This is because the outcome and probabilistic nature of mathematics prove elusive to any detailed human understanding—save by graphical displays, charts, and comparisons between areas of ministerial operations. A fresh perspective needs to be entertained if we are to make a success of our civilizations; despite its adversarial nature, we can make inroads to solutions that benefit everyone.

Table 1. Three-leveled causal hierarchy by Pearl.

| Level | Activity | Questions | Examples |
| --- | --- | --- | --- |
| 1 – Association $P(B\|A)$ | Seeing | What is? How would seeing A change my belief in B? | What does a symptom tell me about a disease? What does a survey tell us about the election results? |
| 2 – Intervention $P(B\|do(A),C)$ | Doing | What if? What if I do A? | What if I take aspirin, will my headache be subsist? What if we ban alcohol? |
| 3 – Counterfactuals $P(B_x\|A',B')$ | Imagination, retrospection | Why? Was it A which caused B? | Was it the aspirin that stopped my headache? What if I had not been drinking the past years? |

References

- Ashby, W. Ross, *An Introduction to Cybernetics*, Chapman and Hall Limited, 1956.

- Beer, A.S. *Management Science*, Aldus Books London, 1956.

- Beer, A.S. *Brain of the Firm*, John Wiley and Sons, 1972.

- Cockshott, W.P., and Cottrell, A. *Towards a New Socialism*, Spokesman, 1993.

- Kant, I. *Critique of Pure Reason*, Konigsberg, 1781.

- Medina, Eden. *Cybernetic revolutionaries: technology and politics in Allende's Chile*, New York, 2014.

- Pearl, J. *Causality,* Cambridge University Press, 2000.

- Spyridopoulos, et. al. "A holistic approach for Cyber Assurance of Critical Infrastructure with the Viable System Model," 10.13140/2.1.2012.3847, 2014.